\documentclass[manuscript]{acmart}
\usepackage[T1]{fontenc}
\usepackage{tabularx}

\acmJournal{TACCESS}
\setcopyright{none}
\acmDOI{}
\acmYear{}
\acmVolume{}
\acmNumber{}
\acmArticle{}
\acmMonth{0}
\title[Beyond Alternative Text]{Beyond Alternative Text:\\ Institutional Semantic Preparation and Local Question Answering\\ for STEM Documents}
\author{Nicola Dall'Acqua}
\orcid{0009-0002-1690-0225}
\affiliation{\department{ETH AI Center}\institution{ETH Zurich}\city{Zurich}\country{Switzerland}}
\email{ndallacqua@ethz.ch}

\author{Gerd Kortemeyer}
\orcid{0000-0001-6643-9428}
\affiliation{\department{Rectorate and ETH AI Center}\institution{ETH Zurich}\city{Zurich}\country{Switzerland}}
\email{kgerd@ethz.ch}

\begin{document}
\begin{abstract}
A scientific graphic can be consulted for its overall pattern, a particular value, or a comparison. For a blind user relying on screen-reader software, a short description may omit the needed detail, while an exhaustive one can turn a simple lookup into lengthy listening. We present a proof-of-concept preparer/reader architecture that separates the amount of information stored from the amount the reader must hear. The institution bears the cost and processing time of using a large, multimodal, reasoning-capable language model to interpret each document once. Preparer software records descriptions, structured data, numerical facts, and source references in a reusable semantic layer. This deliberately verbose layer is intended for machine consumption, not sequential narration to the user: its detail occupies storage rather than listening time. Reader software retrieves relevant evidence from this layer for a small, CPU-based language model to answer the user's questions. The reader targets ordinary consumer laptops and operates without a dedicated GPU or internet connection once the software, models, and document package are installed. Direct document reading remains available, and responsibility for accessible publication and quality assurance remains with the institution. Question answering supplements these obligations rather than replacing them. An initial evaluation of four local models reveals incorrect or incomplete answers and timeouts, showing that prepared evidence alone does not ensure reliable answers. The prototype establishes an inspectable preparation-to-reading workflow; dependable question answering and accessibility benefits remain to be demonstrated.
\end{abstract}

\maketitle
\section{Introduction}
\label{sec:intro}

A scientific graphic can be consulted for its main message, a particular value, the relationship between two categories, or the meaning of an axis. These are different reading tasks. A description that provides good orientation may not contain the detail needed for a numerical comparison, while a lengthy description may make a simple lookup unnecessarily demanding. Research on accessible visualization and context-aware image descriptions has consequently investigated navigable, multimodal, and conversational alternatives to presenting a single fixed account of visual content~\cite{vizability2024,maidr2024,contextaware2024}.

This problem does not diminish the importance of accessible publishing. Text alternatives, meaningful structure, reading order, accessible tables, and mathematical representations remain essential. WCAG and EPUB Accessibility provide requirements for these foundations, and DAISY guidance explicitly accommodates extended descriptions of complex images~\cite{wcag22,epuba11y,daisy}. ``Beyond alternative text'' therefore means supplementing established practice, not suggesting that accessibility standards require every graphic to be reduced to one short paragraph or that a language model can replace conforming content.

Our starting point is a division of responsibility:
\begin{itemize}
\item An author, publisher, or institution should undertake the costly work of preparing and reviewing a reusable document representation.
\item A reader should be able to consult that representation at a chosen level of detail, without having to upload the source repeatedly to a remote model.
\end{itemize}
This is a design position rather than an empirical finding about user preferences. It responds to the potential transfer of preparation work, service costs, and error checking to people seeking access. Studies of AI-based visual assistance make that last concern particularly important: blind users may have to verify or contest plausible but incorrect interpretations using other resources~\cite{misfitting2024,adnin2024}.

We investigate this division through a \emph{preparer} and a \emph{reader}. The preparer interprets a scientific PDF with an external document model and assembles model-authored descriptions, numerical facts, structured rows, and a retrieval index. The reader runs locally, displays ordinary extracted text and prepared object cards, and optionally asks a small language model to formulate an answer from selected evidence. The intended reading condition assumes a consumer CPU and no dedicated GPU.

The central design question is:
\begin{quote}
What can be moved into a durable, institutionally prepared representation so that later interaction requires neither repeated visual interpretation nor a large local model?
\end{quote}
Our prototype investigates that question at the level of an implemented workflow and a small diagnostic case study. It does not yet establish that the resulting interface is accessible in practice or that its question-answering mode is dependable.

The contribution has three parts:
\begin{enumerate}
\item First, we specify a two-stage workflow and a strict cross-file contract that separates the source, model-authored annotations, retrieval records, and presentation surfaces.
\item Second, we demonstrate how a scientific figure can retain both an orienting description and task-relevant numerical annotations in that representation.
\item Third, we report the limitations exposed by a two-document, four-model diagnostic exercise and distinguish preparation, evidence selection, answer formulation, and runtime failures. 
\end{enumerate}
The negative observations are part of the contribution: even when relevant evidence is available, a small local model cannot simply be assumed to function as a faithful verbalizer.

Our proof-of-concept implementation accepts PDF input and exposes an EPUB reading surface, but its semantic resources reside in a custom directory outside the EPUB archive. It has no verified mathematical accessibility pipeline, no demonstrated interoperability with ordinary EPUB readers, and no evaluation with blind or low-vision participants. We use \emph{proof of concept} to describe this implemented but incompletely validated architecture, not to imply an established deployment benefit.

\section{Related Work and Positioning}
\label{sec:related}
Our work draws on research in accessible scientific publishing, AI-assisted information access, and retrieval-augmented generation. These areas address complementary aspects of making complex content accessible, from producing informative descriptions to enabling interactive exploration and supporting trustworthy use of AI-generated responses. In this section, we discuss the relevant work and position our approach in relation to these developments.

\subsection{Interactive access to scientific content}

Accessible publishing already offers more than brief alternative text. EPUB Accessibility sets requirements for accessible publications and for information about their accessibility features, while DAISY provides guidance on extended descriptions of complex images~\cite{epuba11y,daisy}. Such descriptions can explain a figure's organization, labels, values, and relationships. The challenge is to let readers find the information they need without requiring them to work through every available detail.

Earlier work on AI-assisted conversion of \LaTeX{} documents demonstrated that a language model could produce navigable HTML with verbal descriptions of equations, tables, and figures defined in the source code~\cite{kortemeyer2023accessibility}. It also showed that interpretation could come at the expense of faithful reproduction: a concise explanation might omit mathematical indices or other details present in the original. This illustrates both the potential of AI-assisted preparation and the need to distinguish making content easier to read from preserving its full meaning.

Interactive visualization tools address the related problem of letting users choose what to explore. VizAbility combines keyboard navigation with natural-language questions about charts~\cite{vizability2023,vizability2024}. MAIDR provides complementary ways to explore statistical visualizations through text, sound, braille, and review of the data. Its evaluation found that blind participants combined these methods in different ways, depending on their experience and preferences~\cite{maidr2024}. MAIDR Meets AI extends this approach with language-model-generated descriptions and examines how users customize descriptions and check the responses~\cite{maidrai2024}. Together, these studies support offering several ways to examine visual information rather than prescribing a single description or interaction method.

The surrounding document also matters. In a study of AI-generated image descriptions, blind and low-vision participants rated descriptions informed by the surrounding webpage more highly for relevance and quality than descriptions generated from the image alone~\cite{contextaware2024}. Our preparer similarly gives the external model access to the complete paper when describing an individual figure or table. This allows it to consult definitions, explanations, and numerical values reported elsewhere in the document. We do not measure the effect of that context separately.

Our approach combines document preparation with selective access at reading time. The institution undertakes the expensive analysis and remains responsible for providing accessible materials and checking their quality. Detailed descriptions, data, and source references are stored in a reusable semantic layer intended primarily for the local AI system. The blind reader need not listen to that entire layer through screen-reader software; instead, the reader software selects information relevant to the question. The aim is to support this interaction on an ordinary consumer laptop, without a dedicated GPU or internet connection once the software, models, and document package are installed. This complements existing accessible publishing and visualization tools rather than replacing them. Our limited evaluation does not establish benefits across the range of scientific content that the architecture is intended to accommodate.

\subsection{AI assistance, verification, and trust}

Interviews with blind users show that generative AI can be useful while still presenting difficulties with inaccessible interfaces, incorrect answers, and misunderstandings of how the tools work~\cite{adnin2024}. Research on AI-based visual assistance also documents the effort involved in checking results against prior knowledge, other sources of information, or assistance from other people~\cite{misfitting2024}. An answer that sounds helpful may therefore leave substantial work for the user to establish whether it is correct.

Evaluations of multimodal models on tasks informed by visually impaired users identify problems with invented details, multilingual support, and the interpretation of complex visual information~\cite{visualassistants2025}. A diary study likewise found generally favorable participant assessments alongside incorrect answers and cases where the system did not provide an answer~\cite{diary2026}. These findings show why perceived helpfulness, answer correctness, and ease of checking an answer need to be considered separately. They motivate our evaluation but do not establish how often such problems occur in the present reader.

Our prototype retains the source PDF and the prepared descriptions and data. Generated answers identify the document objects from which their supporting information was selected. These references allow users to inspect the associated records, but they do not establish that an answer is correct. Nor do they establish that opening a JSON record or returning to the original PDF is a practical way for a blind reader to check an answer. Whether the available information supports effective verification remains a question for user testing.

\subsection{Retrieval from a prepared semantic layer}

Retrieval-augmented generation first finds relevant information in a stored collection and then supplies it to a language model to help answer a question~\cite{lewis2020}. Our reader follows this approach, but searches prepared descriptions, numerical facts, and data rows rather than passages taken directly from the original document. One search mode uses BGE-M3 embeddings to compare questions and records by their estimated similarity in meaning~\cite{bgem3}. An alternative mode ranks the same records by matching words in the question.

The contribution is not a new search algorithm. It is the preparation and storage of detailed, reusable information between the initial analysis of a document and the user's later questions. The large external model interprets the document during preparation; the small local model receives selected information from that stored representation rather than interpreting the original PDF. The semantic layer can therefore contain considerably more detail than would be appropriate for a description intended to be read from beginning to end.

This division also makes it possible to examine different sources of error. The preparer may record a value incorrectly. The search may fail to find a correctly recorded value, or may select only part of the information needed. Even when the selected information is sufficient, the local model may omit a value or describe a relationship incorrectly. Checking that the package files are well formed and agree with one another does not check any of these aspects of answer quality. The architecture and evaluation below therefore distinguish errors in document preparation, information retrieval, and answer generation.
\section{Implemented Preparer/Reader Architecture}
\label{sec:architecture}
In this section, we present our proposed architecture into a separate preparer and reader. We emphasize the layout of a prepared document, the way in which additional information is stored, and how that is used to answer questions locally. The section concludes with a worked example to illustrate our architecture.

\subsection{Preparing the document at the institution}

The preparer processes one PDF-document at a time. The person preparing the document provides and checks its title and language. The document must contain machine-readable usable text, rather than only images of printed pages. The program keeps an unchanged copy of the source and calculates a digital fingerprint, or hash, that can later be used to check whether the file has changed. It uses PyMuPDF to extract text from each page and saves page images to help investigate problems.

If the extracted text is unusually sparse or appears corrupted, preparation stops. The operator can explicitly allow processing to continue, but the resulting package is marked experimental. The program does not recognize text in scanned images through optical character recognition (OCR), nor does it correct the order in which passages are extracted. Thus, even if the external model can interpret a scanned page, the prototype still needs usable extracted text for ordinary document reading.

The complete PDF is uploaded once to a multimodal model hosted on Azure. The model first identifies the document's sections and individual elements, such as figures, tables, and diagrams. These individually described elements are called \emph{objects} in the software and diagrams. For each element, a separate request then produces a self-contained description (a ``card'' in the diagrams), together with supporting data and source references. The description provides information about that element for the local language model to use when answering questions.

Every request gives the external model access to the complete PDF. It also supplies text extracted from the element's first page and the immediately preceding and following pages. This neighboring text helps locate the element; it does not limit which parts of the paper the model can consult. The initial list of elements is also generated by the model, rather than assembled by a program that searches for captions.

\begin{figure}[t]
\centering
\includegraphics[width=0.68\linewidth]{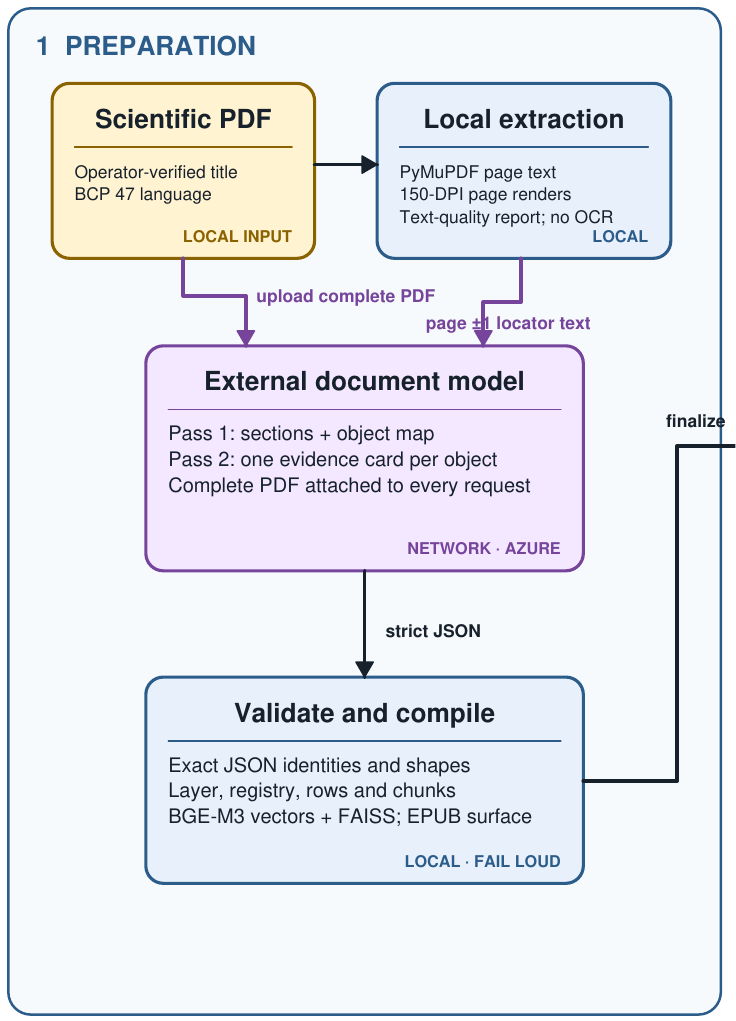}
\caption{Institution-side preparation. The external model interprets the document, while text extraction, checks on the returned data, and creation of the package files take place locally. The outgoing arrow leads to the completed document package. This detail is taken from the overall system diagram; the complete diagrams are provided in the accompanying architecture atlas.}
\Description{An operator supplies a scientific PDF, a checked title, and a language setting. Local software extracts the text and creates page images for investigating problems. The complete PDF is uploaded to an Azure-hosted model, which first identifies sections and document elements and then produces a self-contained description, called a card, for each element. Local software checks the returned data and assembles descriptions, data rows, search resources, and an EPUB reading version. An outgoing arrow represents the completed document package.}
\label{fig:preparation}
\end{figure}

The model returns its results in JSON, a text format that stores information in named fields. The preparer checks the required fields and assembles the semantic layer, the lists of document elements and sections, and the separate data files. It then creates searchable records, their numerical representations for similarity-based search, and the search index. Finally, it produces the EPUB reading version and a package inventory that records the files and preparation settings. An identical copy of this inventory is saved as the preparation report. A file marking the package as incomplete is removed only after all steps have finished. Figure~\ref{fig:preparation} summarizes this process.

When an error occurs, the program stops rather than attempting to conceal or repair it. It does not invent missing fields, repair incorrectly formatted JSON, retry failed requests, switch models, or continue from a partially completed package. Information already saved remains available for investigating the failure, and the package remains marked incomplete. A new preparation run replaces the output folder. This makes errors explicit, but a temporary interruption near the end of processing can require the entire preparation to be repeated.

\subsection{Descriptions and data for each document element}

Each element has an identifier, type, label, title, short summary, and detailed description. Its record also includes element-specific information, data rows, individual numerical facts, source references, a model-reported confidence value, and warnings. Two text fields have different purposes: \texttt{embedding\_text} is used to find records relevant to a question, while \texttt{prompt\_text} contains the information supplied to the local answer model. Separate entries for data rows and numerical facts make individual values searchable rather than leaving them only within a longer prose description.

The file format permits different kinds of document elements rather than limiting them to a fixed list of figure types. This flexibility does not mean that the prototype can interpret and present every kind of scientific content accessibly. It has no dedicated equation parser, conversion to MathML, or system for speaking mathematical notation. MathML is a format for representing the structure of mathematical expressions. The prototype also does not represent the positions of individual table cells or provide specialized checks for chemical or circuit diagrams. Accessible mathematics therefore remains a requirement for further development, not a demonstrated capability.

The external model is asked to preserve labels, units, relationships, reading order, and uncertainty. It is also asked to distinguish values explicitly reported in the source from values estimated by inspecting a graphic. The software checks whether the returned information follows the required format and agrees with the initial document inventory. For example, an element's identifier and first source page must remain consistent between the initial list and the detailed description.

These checks do not compare a reported value, unit, relationship, or visual estimate with the original document. Numerical information within rows and fact entries is not required to follow one common arrangement of fields, and the model's confidence values have not been shown to correspond to measured probabilities of correctness. We therefore treat the results as \emph{model-generated descriptions and data}, rather than independently verified facts.

\subsection{The document package and its EPUB reading version}

The prepared document is distributed as a folder containing several related files. Together, these files form the \emph{package}; the EPUB file alone is not the complete package. The current software requires the files to follow version 1.0 of the project's format and to agree with one another. Table~\ref{tab:package} summarizes their contents.

The file \texttt{book.epub} provides two reading views, written in XHTML, a form of HTML used in EPUB publications. The Normal view contains a document summary, warnings, and the extracted text grouped by PDF page. The Structure view contains the prepared descriptions of individual document elements. The Normal view retains the text's extraction order and spacing rather than reconstructing a hierarchy of sections and subsections.

\begin{table}[t]
\caption{Principal files in the document package. The EPUB contains the two reading views; the other files are stored alongside it in the same package folder.}
\label{tab:package}
\centering
\begin{tabularx}{\linewidth}{@{}p{0.29\linewidth}X@{}}
\toprule
File or resource & Contents \\
\midrule
\texttt{book.epub} & Two XHTML reading views containing extracted text and prepared descriptions, together with EPUB navigation files. Original graphics, structured data rows, and MathML are not included. \\
\path{accessibility-layer.jsonld} & Page text, section information, and model-generated descriptions and data in the project's own JSON format. Despite the filename, this is not an interoperable JSON-LD document. \\
Element lists and row files & Lists of document elements and sections, and separate copies of the data rows associated with each element. \\
Search records and index & One searchable summary per element and a separate record for each numerical fact and data row, with BGE-M3 numerical representations and a FAISS search index. \\
Source and package inventory & An unchanged copy of the PDF, its digital fingerprint, file locations, counts, and preparation settings. The inventory is also saved as the preparation report. \\
\bottomrule
\end{tabularx}
\end{table}

The semantic layer, structured data, and search resources are stored alongside the EPUB, not inside it. Ordinary EPUB software has access to the two reading views but cannot use these additional resources for question answering. The custom reader therefore opens the complete package folder. Although the semantic-layer file is named \texttt{accessibility-layer.jsonld}, it contains ordinary project-specific JSON. It lacks the shared definitions used by JSON-LD to exchange linked data, so the filename should not be taken as evidence of compatibility with other JSON-LD software.

When opening a package, the reader checks the source fingerprint, file locations, element identifiers, and agreement between the different copies of the data. It also checks the order of the search records and their correspondence with the numerical representations and search index. These checks reject incomplete packages and files that have been mixed incorrectly. They do not establish that the model's descriptions or data are accurate: an incorrect value can be stored consistently in every file.

The original PDF remains available for reference. Prepared descriptions include page references and internal links, but the links lead to the descriptions rather than to precisely marked regions of the original figures. The inspected example package contains no coordinates identifying such regions.

\subsection{Reading and asking questions locally}

The browser reader offers three modes. Normal displays the extracted document text. Structure presents the prepared descriptions of individual elements and links to their complete records. Ask accepts questions and displays either a generated answer or the selected supporting information.

The interface uses headings, clearly marked navigation areas, ordinary links, keyboard-operable controls, a link for skipping navigation, and explicit status messages. In Ask mode, Enter submits a question and Shift+Enter inserts a new line. When an answer or error appears, keyboard focus moves to it. Status messages use a screen-reader announcement setting intended not to interrupt ongoing speech. These features are present in the implementation, but they have not yet been evaluated with assistive technologies. One known limitation is that the browser pages identify their language as English regardless of the document's language.

At startup, the reader first checks the selected answer mode. In local-model mode, the separate program running the language model must respond before the reader checks the package, opens its EPUB views, and prepares the search. An unavailable model can therefore prevent startup even when the user only intends to read the text. The reader can be started without answer generation, but this must be chosen explicitly; it does not automatically switch to that mode when the model fails.

\begin{figure}[t]
\centering
\includegraphics[width=0.87\linewidth]{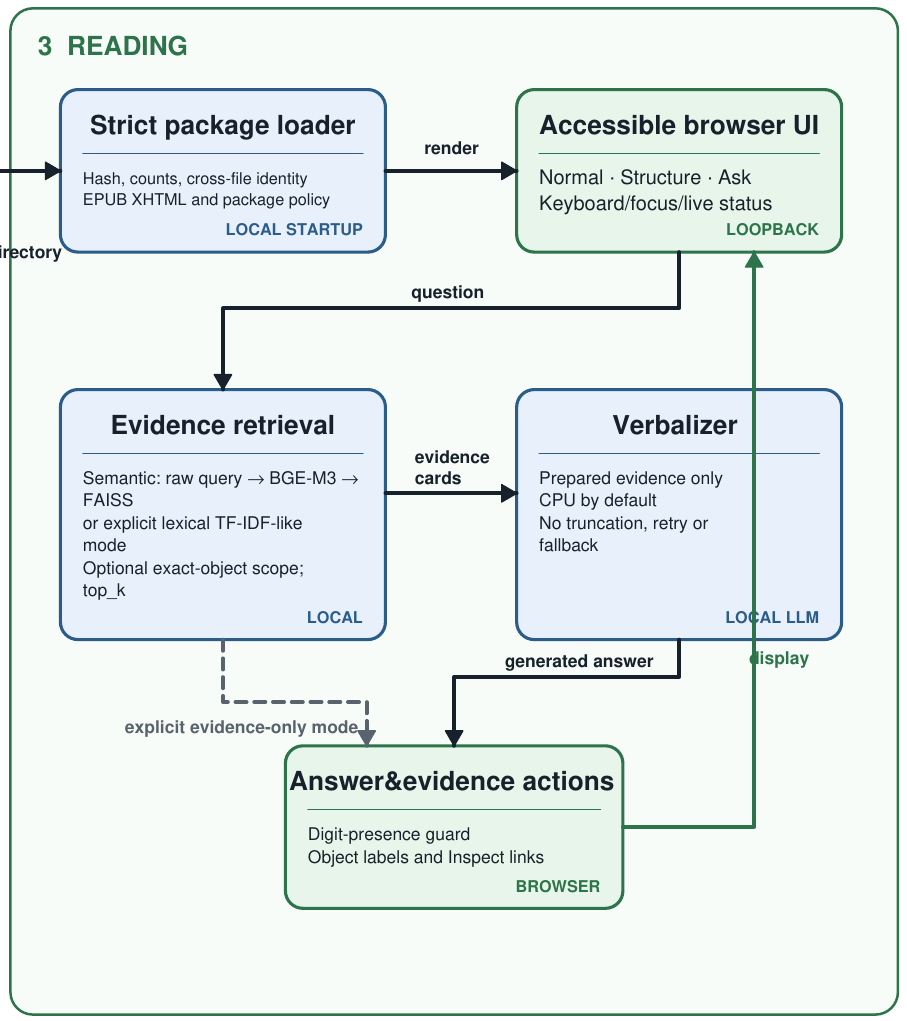}
\caption{Reading and question answering on the user's computer. The incoming arrow represents the complete document package. Meaning-based or word-based search selects prepared information. Depending on the chosen answer mode, this information is sent to a local language model or displayed directly. The reader does not switch modes automatically when generation fails. This detail is taken from the overall system diagram.}
\Description{The complete package enters a loader that checks the source fingerprint, consistency between files, and the EPUB reading views. The browser offers Normal, Structure, and Ask modes. A question is used to search the prepared records by estimated similarity in meaning or by matching words. The search can be restricted to a specifically named document element. Selected information is sent to the local LLM or displayed directly when answer generation is disabled. Generated answers are checked for numbers absent from the supplied information and question. The browser displays the answer, the labels of the source elements, and links for inspecting their complete records.}
\label{fig:reader}
\end{figure}

Each question is handled independently rather than as part of an ongoing conversation (Figure~\ref{fig:reader}). The reader removes extra spaces at the beginning and end but does not rewrite or translate the question. Explicit references to document elements are matched against stored identifiers, labels, or titles after standardizing capitalization, punctuation, and equivalent character forms. Ambiguous references, or a disagreement between a named element and an explicitly selected search restriction, produce an error rather than a guess.

The reader provides two alternative search methods. Meaning-based search uses BGE-M3 to represent the question and prepared search text as lists of numbers, called embeddings. FAISS ranks these representations by similarity, using the inner product after scaling each vector to unit length. The alternative word-based search ranks records by matching terms, giving greater weight to terms that are less common in the collection. This uses a score related to term frequency--inverse document frequency (TF-IDF) and recognizes letters and numbers from different writing systems. The two methods are selected separately; their rankings are not combined. When the question is restricted to particular document elements, only records belonging to those elements are retained after ranking.

The setting \texttt{top\_k} controls how many records are selected, with a default of six. These records are not passages of equal length. A summary may contain a long description, whereas a numerical fact may occupy only a few lines. Each element contributes one summary record and a separate record for each fact and data row. The original page text and section summaries are not included in this search collection.

The selection step does not check whether the combined text will fit within the answer model's input limit, reconsider the ranking with a second method, or verify that all information needed for the question has been found. Meaning-based search has no minimum relevance score, so even weak matches can be selected. The reader also has no dedicated operations for querying fields in structured rows or performing arithmetic. Selecting fewer records therefore does not necessarily produce either a sufficiently informative or a sufficiently short input.

The local answer model receives the question and the selected records' element identifiers, labels, and \texttt{prompt\_text}. It does not receive the PDF. Source pages, warnings, and model-reported confidence reach it only if that information is also included in the selected text. Consequently, information retained elsewhere in the package is not necessarily available when an answer is generated.

The default local setup uses a local LLM, for example the 1.5-billion-parameter Apertus model~\cite{apertus2026}, on the CPU --- a GPU would be used if present and compatible, but we do not expect it. Random sampling is disabled, and each answer is limited to 512 tokens, the text units processed by the model. Before generating an answer, the model service checks whether the complete input, together with the reserved space for the answer, fits within its capacity. If not, it rejects the request rather than deleting part of the input.

A simple numerical check rejects generated answers containing digit-based numbers that occur neither in the question nor in the supplied information. This can detect some newly introduced numbers, but it does not check their meaning. A number already present in the evidence can still be assigned to the wrong category, given the wrong unit, or used in an incorrect comparison.

Answers identify the document elements from which information was selected and provide links to inspect their complete prepared records. They do not link each individual statement to the particular passage or data entry supporting it. When generation is explicitly disabled, the reader instead displays the selected prepared information directly. This still consists of descriptions and data generated during preparation; it is not a comparison condition based on unmodified text from the original document.

\subsection{A worked example: preparing and accessing a scientific paper}
\label{sec:worked-example}

To illustrate how the proposed application handles a scientific document, this section follows the preparation and subsequent use of the article \emph{Observation of Gravitational Waves from a Binary Black Hole Merger} by Abbott et al.~\cite{obsgrav2016}, to which we refer as the \emph{Gravitational Waves} article. The document reports the first direct detection of gravitational waves and the observation of a binary black hole merger. It is a useful example because it contains several types of scientific content that pose different accessibility challenges, including multi-panel figures, numerical tables, mathematical equations, and technical descriptions.

The example is presented in three stages:
\begin{enumerate}
    \item \textbf{Preparation:} the structured descriptions, numerical information, and other accessibility data generated by the preparer.
    \item \textbf{Retrieval:} the information selected from the prepared package and supplied to the local language model when a reader asks a question.
    \item \textbf{Response:} the answer generated by the local model from the retrieved evidence.
\end{enumerate}

\subsubsection{What the preparer produces}

The original article is a 16-page research letter. The preparer, in this case using a \emph{GPT-5.6 Sol} deployment, produces an EPUB3 document along with a structured accessibility layer and supporting data files. In this example, the preparation process identified and analyzed 14 document objects and organized the article into 15 sections. The resulting package contains descriptions and structured representations of figures, equations, and tables, as well as publication metadata, references, and collaboration information.

Table~\ref{tab:prepared-package-overview} summarizes the main types of information extracted from the article. The object counts refer to the semantic objects identified in the prepared document; they do not represent the number of files in the package.

\begin{table}[t]
\centering
\small
\caption{Overview of the 14 semantic objects identified in the prepared gravitational waves article.}
\label{tab:prepared-package-overview}
\begin{tabularx}{\linewidth}{@{}l r X@{}}
\toprule
\textbf{Object type} & \textbf{Count} & \textbf{Examples of prepared information} \\
\midrule
Figures and diagrams & 4 & Panel structure, visual elements, axes, trends, legends, and relationships between plotted quantities. \\
Equations & 4 & Mathematical expressions, symbol definitions, units, and explanations of their application in the article. \\
Parameter tables & 1 & Parameter names, reported values, uncertainties, units, and relevant qualifications. \\
Publication metadata & 1 & Article title, authorship, journal, publication date, and DOI. \\
Bibliography & 1 & The 118 numbered references and their bibliographic information. \\
Author and affiliation information & 3 & Collaboration byline, institutional affiliations, and author notes. \\
\bottomrule
\end{tabularx}
\end{table}

Descriptions go beyond simply extracting the text surrounding an object. They aim to make the information conveyed by the visual or mathematical representation explicit. Three examples illustrate this distinction.

\paragraph{Figures and diagrams.}
The four figures of the gravitational waves article are summarized in Table \ref{tab:prepared-figure-examples} and comprise the detector observations in Figure~1, the physical evolution of the merger in Figure~2, the Advanced LIGO detector diagram in Figure~3, and the search results in Figure~4. Their descriptions capture information such as panel layout, reading order, axes, plotted quantities, and relationships between visual elements. Note that the numbering of the figures correspond to those of the original document~\cite{obsgrav2016}.

For example, the description of Figure~1 explicitly distinguishes the two detector columns: Hanford (H1) is shown on the left and Livingston (L1) on the right. It describes the four rows of panels that cover the observed strain, waveform comparisons, residuals, and time-frequency representations. It also records the relevant relationship between the detector signals: Gravitational waves were detected by Livingston first and at Hanford approximately 6.9 milliseconds later. The description further explains that the Hanford trace is shifted in time and inverted to account for the detectors' relative orientations. These details are important for interpreting the figure rather than simply identifying its subject.

Figure~2 provides another example. Its description distinguishes the upper waveform plot from the lower plot containing two vertical axes. In particular, the green curve is identified as the effective relative velocity, read against the left axis, and the black curve as the effective separation, read against the right axis. It also describes their opposing trends as the merger approaches: the relative velocity increases while the separation decreases. Approximate endpoint values are included as visual estimates rather than as exact tabulated measurements.

\begin{table}[t]
\centering
\small
\caption{Examples of information made explicit in the prepared descriptions of the gravitational waves article's figures.}
\label{tab:prepared-figure-examples}
\begin{tabularx}{\textwidth}{l l X}
\toprule
\textbf{Source} & \textbf{Content represented} & \textbf{Overview of information captured by the preparer} \\
\midrule
Figure~1, p.~2 & Detector observations & Eight-panel layout, detector identities, arrival order, 6.9\,ms delay, time shift, sign inversion, waveform reconstructions, and upward chirp. \\
Figure~2, p.~3 & Merger evolution & Inspiral, merger and ringdown stages; waveform evolution; axis assignments; decreasing separation and increasing relative velocity. \\
Figure~3, p.~4 & Detector design & Orthogonal 4\,km interferometer arms, detector locations, and noise spectra. \\
Figure~4, p.~6 & Search results & Candidate-event distributions, noise backgrounds, and the reported detection significances of the two searches. \\
\bottomrule
\end{tabularx}
\end{table}

\paragraph{Tables and numerical information.}
The preparer also converts the article's parameter table into structured records, which are presented in Table \ref{tab:prepared-source-parameters}. Table~I of the original document reports six source parameters. Rather than retaining only a textual description of the table, the prepared representation stores each parameter together with its reported median, upper and lower uncertainties, unit, and relevant qualifications.

For instance, the primary black hole mass is represented as a source-frame mass of \(36\,M_\odot\), with an upper uncertainty of \(5\,M_\odot\) and a lower uncertainty of \(4\,M_\odot\). The distinction between the two uncertainties is preserved, as is the fact that the reported masses are in the source frame. The prepared data also records that the values are posterior medians with 90\% credible intervals, including statistical and systematic errors. These qualifications matter when interpreting or comparing the numerical values.

\begin{table}[t]
\centering
\small
\caption{Selected entries from the prepared representation of Table~I in the gravitational waves article.}
\label{tab:prepared-source-parameters}
\begin{tabular}{@{}l c c@{}}
\toprule
\textbf{Parameter} & \textbf{Median} & \textbf{90\% credible interval} \\
\midrule
Primary black hole mass & \(36\,M_\odot\) & \(+5/-4\,M_\odot\) \\
Secondary black hole mass & \(29\,M_\odot\) & \(+4/-4\,M_\odot\) \\
Final black hole mass & \(62\,M_\odot\) & \(+4/-4\,M_\odot\) \\
Final black hole spin & \(0.67\) & \(+0.05/-0.07\) \\
Luminosity distance & \(410\,\mathrm{Mpc}\) & \(+160/-180\,\mathrm{Mpc}\) \\
Source redshift & \(0.09\) & \(+0.03/-0.04\) \\
\bottomrule
\end{tabular}
\end{table}

\paragraph{Equations and mathematical relationships.}
Four equations were identified and prepared with additional explanatory information. For each, the package contains the symbolic expression, a textual representation, definitions of the symbols, and an explanation of how the equation is used in the article. For example, the chirp-mass equation
\[
\mathcal{M} = \frac{(m_1 m_2)^{3/5}}{(m_1 + m_2)^{1/5}} = \frac{c^3}{G} \left[\frac{5}{96}\pi^{-8/3}f^{-11/3}\dot{f}\right]^{3/5},
\]
in Section~II of the original document is accompanied by definitions of the component masses, the gravitational-wave frequency, its time derivative, and the physical constants. The prepared description also connects the equation to its application in the paper: the observed frequency evolution gives a chirp mass of approximately \(30\,M_\odot\), implying a detector-frame total mass of at least approximately \(70\,M_\odot\).

\paragraph{Document structure and supplementary information.}
Beyond these scientific objects, the preparer creates a structured representation of the article's 15 sections, including section headings, page ranges, summaries, and links to associated objects. It also extracts publication metadata, the bibliography, the collaboration's author list, institutional affiliations, and author notes. These elements provide navigational and contextual information that is not necessarily central to a question about the scientific results but are still essential when cross-referencing and exploring the document.

Overall, the prepared package contains 14 analyzed objects, 167 structured data rows, and 252 retrieval chunks. The structured rows include, among other information, the six parameter entries in Table \ref{tab:prepared-source-parameters}, panel-level descriptions of the figures of the article, and the 133 institutional affiliations. The retrieval chunks provide smaller units of textual evidence for the subsequent retrieval stage. The package therefore combines a conventional EPUB3 reading surface with a more detailed and machine-readable representation of the document. The next stage examines which parts of this representation are actually selected when a user asks the local model a question.

\subsubsection{What retrieval sources.}
\begin{figure}[t]
\centering
\includegraphics[width=0.87\linewidth]{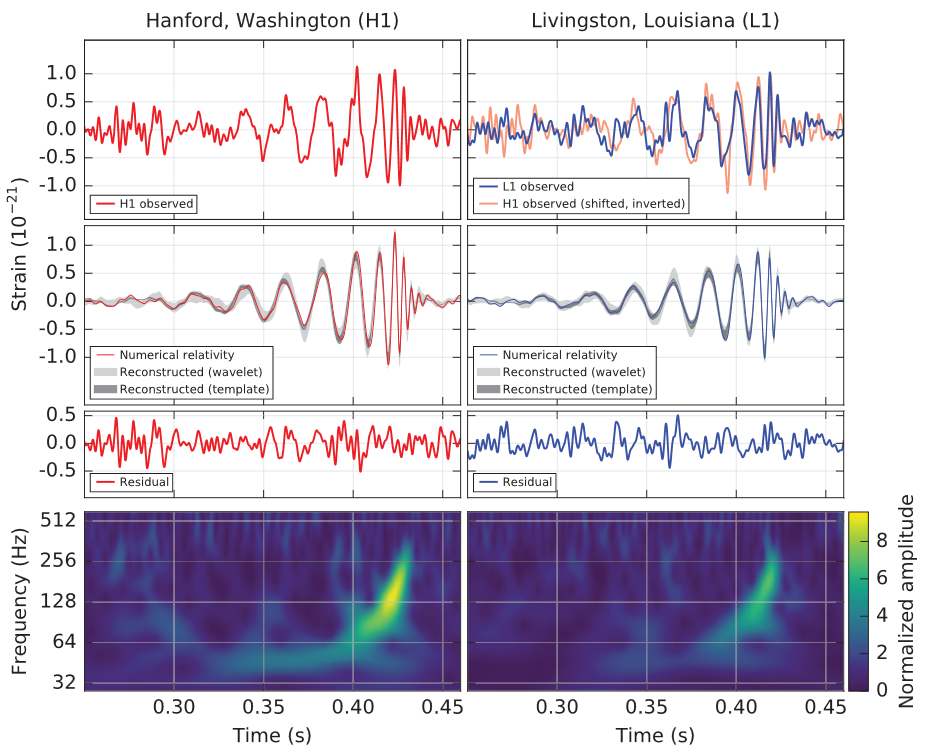}
\caption{Figure 1 of the gravitational waves article~\cite{obsgrav2016}.}
\Description{Figure 1 of the article \emph{Observation of Gravitational Waves from a Binary Black Hole Merger} by Abbott et al.~\cite{obsgrav2016} displays 8 panels arranged in 4 rows and 2 columns. One column per detector (Hanford H1 and Livingston L1). Row one features the observed filtered strain of both detectors, row two the numerical relativity with 90\% credible regions, row three shows residuals and row four a time-frequency representation of the strain-data. For more information we refer to the original article.}
\label{fig:obsgrav_fig1}
\end{figure}

To illustrate how the prepared information is used at query time, consider the question ``What does Figure 1 display?'', which we display, for reference, in Figure \ref{fig:obsgrav_fig1}. The retrieval component selects the six highest-ranked chunks of the prepared document, using the \emph{BGE-M3} meaning-based search mode. All six chunks originate from the prepared evidence object associated with Figure 1. Four of them describe the individual panel rows of the figure, while the remaining two provide numerical facts extracted from the figure caption.

Table~\ref{tab:figure1-retrieval} summarizes the six evidence cards retrieved and the information they provide.

\begin{table}[ht]
\centering
\small
\caption{The six evidence cards retrieved for the question ``What does Figure 1 display?''}
\label{tab:figure1-retrieval}
\begin{tabular}{p{0.19\linewidth} p{0.73\linewidth}}
\hline
\textbf{Evidence card} & \textbf{Retrieved information} \\
\hline
Reading order 1 (top row) &
Hanford H1: Observed filtered strain, red line. Livingston L1: Observed filtered strain, blue line, overlaid with the H1 observation shifted by 6.9 ms and inverted. The aligned traces closely coincide. \\[0.5em]

Reading order 2 (second row) &
Solid numerical-relativity waveforms, dark-gray templates, and light-gray wavelet 90\% credible regions are shown for both detectors. The waveform oscillations increase in frequency and amplitude towards a peak and then decay; both reconstruction methods track the numerical-relativity waveform. \\[0.5em]

Reading order 3 (third row) &
Residual signals after subtracting the filtered numerical-relativity waveform: red for Hanford H1 and blue for Livingston L1. These panels display what remains in each detector after model subtraction. \\[0.5em]

Reading order 4 (bottom row) &
Time-frequency maps for both detectors, with normalized amplitude encoded by colour. Both maps show an upward chirp, indicating increasing signal frequency over time. \\[0.5em]

Numeric fact 1 &
The numerical-relativity waveform was independently confirmed to a level of 99.9\%. Exact value stated in the figure caption. \\[0.5em]

Numeric fact 2 &
The upper cutoff frequency of the visualization bandpass filter applied to all displayed time series is 350 Hz. Exact value stated in the figure caption. \\
\hline
\end{tabular}
\end{table}

The retrieved evidence thus provides descriptions of all four panel rows, including the relationship between the two detectors, as well as two additional quantitative details. As such both information on orientation within the figure and task-relevant numerical annotations are present.
Rather than requiring the model to interpret the original figure directly, retrieval provides these prepared textual descriptions in a structured form. The model's input for this query consists of the user question, the six evidence cards, and a system prompt requesting the model to textually prepare the evidence.

\subsubsection{What the model responds.}

For this example, \emph{Qwen/Qwen3.5-0.8B} was used as the local model.
Based on the question and the evidence retrieved, the model generates the following response.

\begin{quote}
Figure 1 displays observations from two detectors (Hanford H1 and Livingston L1) in three different reading orders. The content includes:
\begin{itemize}
    \item \textbf{Reading Order 1:} Observed filtered strain, red line.
    \item \textbf{Reading Order 2:} Solid numerical-relativity waveform with dark-gray template and light-gray wavelet 90\% credible regions.
    \item \textbf{Reading Order 3:} Red residual after subtracting the filtered numerical-relativity waveform.
    \item \textbf{Reading Order 4:} Time-frequency map with normalized amplitude encoded by color.
\end{itemize}
Additionally, the figure contains specific numeric facts regarding the visualization bandpass filter (upper cutoff of 350 Hz) and the confirmation level of the numerical-relativity waveform (99.9 percent).
\end{quote}

The response accurately captures the main content of the retrieved evidence (see also Figure \ref{fig:obsgrav_fig1}), however, it contains a counting error and omits some details. 
It identifies the four types of panels and their orientations, and includes both numerical facts. However, the introductory sentence incorrectly refers to ``three different reading orders'', whereas the subsequent list correctly describes four. The response also omits some detector-specific details present in the retrieved evidence. For example, it does not mention the 6.9 ms shift and inversion of the Hanford trace, and its description of the residuals refers only to the red Hanford residual, omitting the blue Livingston residual. These omissions do not invalidate the answer to the broad question, but they illustrate that the model does not necessarily preserve all details supplied by retrieval.

\subsubsection{A gap between evidence and model response.}
\begin{figure}[t]
\centering
\includegraphics[width=0.5\linewidth]{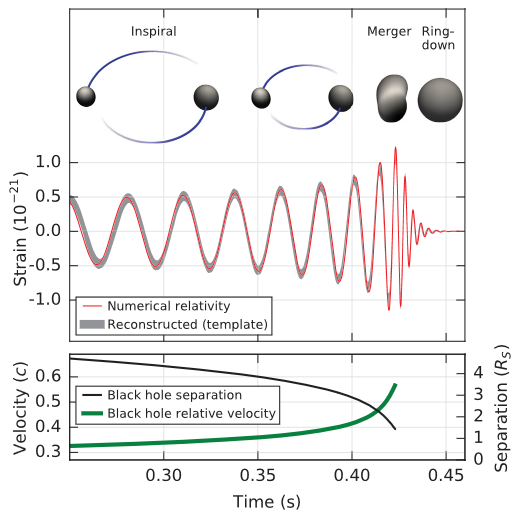}
\caption{Figure 2 of the gravitational waves article~\cite{obsgrav2016}, which shows black hole separation and relative velocity.}
\Description{Figure 2 of the article \emph{Observation of Gravitational Waves from a Binary Black Hole Merger} by Abbott et al.~\cite{obsgrav2016} displays the estimated gravitational-wave strain amplitude projected onto H1 in the top row. The bottom row displays black hole separation and their relative velocities, where separation uses the right-hand vertical axis and velocity the left-hand vertical axis.}
\label{fig:obsgrav_fig2}
\end{figure}

To further illustrate this observed gap between what retrieval sources and what the local model verbalizes, we use \emph{HuggingFaceTB/SmolLM2-1.7B-Instruct} and the question ``In the plot of black-hole separation and relative velocity, which quantity uses the left-hand vertical axis and which uses the right-hand vertical axis and how does each quantity change as the merger approaches?''. The relevant plot is shown in Figure \ref{fig:obsgrav_fig2}. This question deliberately ignores the top row of the plot and focuses only on the bottom row, mimicking real-world reading tasks.

For the sake of brevity, we only provide the two relevant of the six total retrieved evidence cards, these can be found in Table \ref{tab:figure2-retrieval}. The model generated the following response.

\begin{quote}
    The quantity that uses the left-hand vertical axis is the effective black-hole separation ($R_S$) and the quantity that uses the right-hand vertical axis is the effective relative velocity (v/c). As the merger approaches, the effective black-hole separation decreases and the effective relative velocity increases.
\end{quote}
By comparing this with the relevant evidence cards and the original plot, it is evident that the local model reversed axis assignments, incorrectly claiming that separation uses the left-hand vertical axis and velocity the right-hand vertical axis. Despite being provided with the correct assignments in the evidence, the local model confused the two. Nevertheless, it did correctly verbalize the change of both quantities as the merger approaches.

This serves as further evidence that local models do not necessarily produce correct responses from correct retrieved evidence. More details on model response quality can be found in Section \ref{sec:method}.

\begin{table}[ht]
\centering
\small
\caption{The relevant evidence card retrieved for the question ``In the plot of black-hole separation and relative velocity, which quantity uses the left-hand vertical axis and which uses the right-hand vertical axis and how does each quantity change as the merger approaches?''}
\label{tab:figure2-retrieval}
\begin{tabular}{p{0.19\linewidth} p{0.73\linewidth}}
\hline
\textbf{Evidence card} & \textbf{Retrieved information} \\
\hline
Reading order 5 (Bottom panel) &
Black hole relative velocity, Left axis, Velocity (c), ranging from approximately 0.32 c to approximately 0.56 c. Effective relative velocity increases as merger approaches. \\[0.5em]

Reading order 4 (Bottom panel) &
Black hole separation, Right axis, Separation ($R_S$) ranging from approximately 4.5 $R_S$ to approximately 1.4 $R_S$. Effective separation decreases as merger approaches. \\[0.5em]

$\ldots$ & $\ldots$ \\

\hline
\end{tabular}
\end{table}

\subsection{Computing requirements and use of external services}

The software and local models must be installed in advance, and the prepared document package must be available on the user's computer. After this setup, the intended local reader can answer questions without an internet connection. Its components communicate within the same computer, and the models are loaded from locally stored files.

Document preparation, however, sends the complete PDF to the external service. Deleting the uploaded file is an optional setting and is not enabled by default. The institution therefore needs to review permission to upload the material and the service's retention arrangements separately. Keeping reading-time questions local does not remove these responsibilities or establish that every aspect of document handling is private.

Using the computer's main processor avoids requiring a dedicated graphics processor, but it does not guarantee low memory use, low energy consumption, or acceptable response times on every consumer laptop. In the meaning-based search and local-answer configuration, both models remain loaded in memory after startup.

The local model generates one answer at a time. The reader stops waiting after 180 seconds and reports a timeout, but this does not cancel the model's work. A timed-out request can therefore continue consuming processing time and delay subsequent questions.

\subsection{AI Usage}

For the implementation of the preparer and the reader, artificial intelligence was used as a coding agent (Codex, OpenAI GPT versions 5.x and 6.x). ChatGPT (OpenAI) was used in author-directed, iterative discussions during the conceptual development and refinement of the preparer/reader architecture. These discussions addressed the division between institutional document preparation and local reading, the organization of the reusable semantic layer, and strategies for selecting and presenting evidence to small local language models. ChatGPT contributed design suggestions, comparisons of alternatives, and analysis of limitations and failure modes encountered during development. The research objectives and practical constraints were specified by the authors, who selected and revised the proposals incorporated into the work and retain responsibility for the resulting design and its presentation.
\section{Proof-of-Concept Evaluation}
\label{sec:method}
To evaluate the feasibility of our proof of concept, a descriptive analysis is performed within the reader. To that end, two pre-prepared packages alongside a set of questions and reference answers were utilized to evaluate the capability of small local models at verbalizing evidence.

\subsection{Evidence and scope}

For the proof of concept, two prepared packages were used. First, a 29-page version of \emph{Multilingual Performance of a Multimodal Artificial Intelligence System on Multisubject Physics Concept Inventories}~\cite{kortemeyer2025}, which we refer to as the \emph{Multilingual Performance} article. Second, a 16-page version of the previously utilized \emph{Gravitational Waves} article~\cite{obsgrav2016} is used. 
In addition, four locally executable language models were considered: \path{Qwen/Qwen3.5-0.8B}, \path{LiquidAI/LFM2.5-1.2B-Instruct}, \path{HuggingFaceTB/SmolLM2-1.7B-Instruct}, and \path{swiss-ai/Apertus-v1.1-1.5B-Instruct}, which, throughout this work, will be referred to as \emph{Qwen}, \emph{LFM}, \emph{SmolLM}, and \emph{Apertus}. For each paper, a set of questions was designed to assess the reader's ability to retrieve and synthesize information from the prepared document, ranging from closed-ended questions, such as extracting numerical values from figures, to open-ended questions, such as summarizing the limitations discussed in the paper. All experiments were conducted on an AMD Ryzen 7 5700U CPU without a dedicated GPU, reflecting realistic deployment conditions in which users cannot be assumed to have access to powerful hardware. To ensure consistent evaluation conditions, all models were run with a context window of 4,096 tokens and a deterministic output limit of 512 tokens.

\subsection{Evaluation questions and diagnostic conditions}

For each of the two prepared packages, nine diagnostic conditions were defined, covering four question families. The questions for the multilingual performance package concern the evaluation of image requirements, subject-category performance, AI-versus-student comparisons, and paper limitations, these can be found in Table~\ref{tab:conditions-paper1}. Specifically, C1--C3 assess overall image-requirement scores under three evidence-selection settings. C4--C6 identify the highest and lowest English subject-category averages, while C7--C8 compare AI and student performance. C9 asks for an overview of the paper's limitations. The numbering of the figures within the scope of evidence selection are consistent with the numbering in the original multilingual performance article.

For the gravitational waves package, the questions concern the gravitational-wave event GW150914 and the associated analysis reported by Abbott et al. (2016) and may be found in Table~\ref{tab:conditions-paper2}. C1--C3 ask for the source-frame masses of the primary, secondary, and final black holes, varying the retrieval depth from six chunks to three and one chunk, respectively. C4--C6 examine the plot of black-hole separation and relative velocity: C4 uses global semantic retrieval, C5 restricts the evidence to Figure~2 and its caption and associated data and semantic layers, and C6 provides only minimal two-sentence evidence with retrieval bypassed. C7--C8 compare the Hanford and Livingston detector observations, including signal arrival order, time delay, and strain-trace alignment; C8 restricts the evidence to Figure~1 and its associated caption, data, and semantic layers, using the top retrieved chunk. Finally, C9 asks for a summary of the three tests of consistency with general relativity described in Section~VI, ``Source Discussion,'' and their overall conclusion. Once more, the numberings are consistent with those in the original gravitational waves article.

Reference answers for both packages are provided in Appendix~\ref{app:questions}. The conditions are not nine independent questions or a factorial experiment. Several conditions vary both object scope and retrieval depth, while C6 in both papers changes the evidence-delivery mechanism.

These questions do not measure accessibility, comprehension, task completion, independence, preference, or trust. Future participatory evaluation is necessary to address those outcomes.

\begin{table}[t]
\caption{Diagnostic conditions for the multilingual performance package.}
\label{tab:conditions-paper1}
\centering
\begin{tabularx}{\linewidth}{l X X}
\toprule
ID & Question & Evidence selection \\
\midrule
C1 & What are the overall text-only, unneeded-image, and required-image scores? & Global semantic retrieval, top 6 \\[0.5em]
C2 & Same as C1 & Figure 6 scope, top 3 \\[0.5em]
C3 & Same as C1 & Figure 6 scope, top 1 \\[0.5em]
C4 & What are the highest and lowest English subject-category averages? & Global semantic retrieval, top 6 \\[0.5em]
C5 & Same as C4 & Figure 3 scope, top 6 \\[0.5em]
C6 & Same as C4 & Retrieval bypassed, minimal two-sentence evidence: ``From highest to lowest average, the categories are Thermodynamics (THERM), Astronomy (AST), Reasoning (REAS), Relativity (RELA), Mathematics (MATH), Mechanics (MECH), Quantum Physics (QP), Electricity and Magnetism—Fields (EM-F), Optics (OPT), Electricity and Magnetism—Circuits (EM-C), and Laboratory Skills (LAB). The surrounding text reports exact averages of 85.2\% for Thermodynamics, 80.4\% for Astronomy, 78.7\% for Reasoning, 75.0\% for Relativity, and 35.0\% for Laboratory Skills.'' \\[0.5em]
C7 & Please give an AI-versus-student comparison. & Global semantic retrieval, top 6 \\[0.5em]
C8 & Same as C7 & Figure 5 scope, top 1 \\[0.5em]
C9 & Please provide an overview of the limitations. & Global semantic retrieval, top 6 \\[0.5em]
\bottomrule
\end{tabularx}
\end{table}

\begin{table}[t]
\caption{Diagnostic conditions for the gravitational waves package.}
\label{tab:conditions-paper2}
\centering
\begin{tabularx}{\linewidth}{l X X}
\toprule
ID & Question & Evidence selection \\
\midrule
C1 & What are the reported source-frame masses of the primary black hole, the secondary black hole, and the final black hole in GW150914? Give the central estimates in solar masses. & Global semantic retrieval, top 6 \\[0.5em]
C2 & Same as C1 & Global semantic retrieval, top 3 \\[0.5em]
C3 & Same as C1 & Global semantic retrieval, top 1 \\[0.5em]
C4 & In the plot of black-hole separation and relative velocity, which quantity uses the left-hand vertical axis and which uses the right-hand vertical axis, and how does each quantity change as the merger approaches? & Global semantic retrieval, top 6 \\[0.5em]
C5 & Same as C4 & Figure 2 scope, top 6 \\[0.5em]
C6 & Same as C4 & Retrieval bypassed, minimal two-sentence evidence: ``Figure 2 illustrates the three stages of a black-hole merger—inspiral, merger, and ringdown—along a shared time axis ranging from 0.30 to 0.45 seconds, the left-hand axis displays black hole separation and the right-hand axis relative velocity. The illustrations show two black holes orbiting and approaching each other, merging into a distorted horizon, and finally relaxing into a single stationary black hole.'' \\[0.5em]
C7 & Please compare the GW150914 observations at Hanford and Livingston: which detector recorded the signal first, what was the delay, and how were the strain traces aligned for comparison? & Global semantic retrieval, top 6 \\[0.5em]
C8 & Same as C7 & Figure 1 scope, top 1 \\[0.5em]
C9 & Please summarize the three tests of consistency with general relativity described in Section VI, 'Source Discussion,' and state their overall conclusion. & Section VI, ``Source Discussion,'' global semantic retrieval, top 6 \\[0.5em]
\bottomrule
\end{tabularx}
\end{table}

\subsection{Scoring, timing, and descriptive analysis}

Each of the four local models was presented with all the questions for both packages and the responses were individually evaluated. Response times were recorded for each question; response times exceeding 180 seconds were considered timeouts. Furthermore, a numerical guard was used on the model responses to prevent the introduction of numerical values that are absent from the question and the prepared evidence. For answer quality, a three-level scoring scheme was applied based on reference answers and acceptance criteria defined in the Appendix~\ref{app:questions}. Fully correct answers received 2 points, partially correct answers received 1 point, and incorrect answers or errors, including timeouts, received 0 points. Partial credit was awarded whenever a response contained some, but not all, of the information required for full credit, as specified in the appendix.

The resulting scores and response times are presented in Section \ref{sec:results} and are intended only for descriptive analysis. This evaluation does not aim to establish a general-purpose benchmark or provide broadly generalizable rankings of the selected models. Rather, it serves to assess their relative suitability within the specific application context, namely answering questions about prepared documents using a locally running model under constrained hardware and inference conditions. The findings should therefore be interpreted in terms of the requirements of this particular application rather than as general statements about model capabilities.

\subsection{AI Usage}

For the proof-of-concept, artificial intelligence (Codex, OpenAI GPT versions 5.x) was used in the creation of a script to efficiently present all local models with all questions for both packages. Furthermore, ChatGPT (OpenAI) was used to create reference answers and acceptance criteria (see Appendix~\ref{app:questions}). The script, the reference answers and the acceptance criteria were validated by the authors, who retain full responsibility.

\section{Results}
\label{sec:results}
Evaluation of the four local reader models returned variable results both across models and across packages. Furthermore, a gap between retrieved evidence and model responses was observed, implying that relevant evidence does not guarantee a faithful response.

\subsection{Recorded scores and operational failures}

\begin{table}[t]
\caption{CPU-only observations for the first package~\cite{kortemeyer2025}. Latencies are approximate ranges for non-timeout responses. Failure counts are based on the recorded execution outcomes. Each model has nine conditions; all failures received zero points.}
\label{tab:results-paper1}
\centering
\begin{tabular}{@{}lrrrrr@{}}
\toprule
Model & Score / 18 & Time (s) & Timeout & Guard & Context \\
\midrule
Apertus 1.5B & 4 & 60--90 & 1 & 0 & 1 \\
Qwen3.5 0.8B & 8 & 70--100 & 3 & 1 & 0 \\
SmolLM2-1.7B-Instruct & 13 & 90--120 & 1 & 1 & 0 \\
LFM2.5-1.2B-Instruct & 5 & 30--60 & 0 & 0 & 0 \\
\bottomrule
\end{tabular}
\end{table}

The totals recorded for the multilingual performance package~\cite{kortemeyer2025} range from 4 to 13 out of 18 (Table~\ref{tab:results-paper1}). SmolLM obtains the highest score, with 13 points, while the other models achieve between 4 and 8 points. Across the 36 model-condition records, the outcome annotations identify five timeouts, two numeric-guard rejections, and one context overflow. All recorded failures received zero points. These counts describe the results of this diagnostic exercise and should not be interpreted as general failure rates. Appendix~\ref{app:scores} provides the scores per condition.

Although SmolLM achieves the highest recorded total, this does not establish its general superiority. The evaluation set is small and development-oriented, several conditions share the same question, and the score combines answer quality with runtime failures. LFM has the shortest completed-response latency range and does not experience recorded operational failures, although its response quality remains comparatively low.

\begin{table}[t]
\caption{CPU-only observations for the second package~\cite{obsgrav2016}. Latencies are approximate ranges for non-timeout responses. Failure counts are based on the recorded execution outcomes. Each model has nine conditions; all failures received zero points.}
\label{tab:results-paper2}
\centering
\begin{tabular}{@{}lrrrrr@{}}
\toprule
Model & Score / 18 & Time (s) & Timeout & Guard & Context \\
\midrule
Apertus 1.5B & 5 & 40--60 & 0 & 2 & 0 \\
Qwen3.5 0.8B & 8 & 60--90 & 0 & 3 & 0 \\
SmolLM2-1.7B-Instruct & 7 & 60--90 & 2 & 2 & 0 \\
LFM2.5-1.2B-Instruct & 12 & 30--60 & 0 & 0 & 0 \\
\bottomrule
\end{tabular}
\end{table}

For the gravitational wave package~\cite{obsgrav2016}, the recorded totals range from 5 to 12 of 18 (Table~\ref{tab:results-paper2}). LFM achieves the highest score, with 12 points, followed by Qwen with 8 points. Across the 36 model-condition records, seven numeric-guard rejections and two timeouts were recorded, with no context overflows. Appendix~\ref{app:scores} provides the scores by condition. SmolLM, which achieved the highest score for the multilingual performance package, performs less favorably here, with two timeouts and a total of 7 points. This variation illustrates that the relative performance of the models depends on the document and the set of questions, further emphasizing the application-specific and descriptive nature of the evaluation.

\subsection{Relevant evidence does not guarantee a faithful answer}

A recurring observation in both prepared packages is that retrieving relevant evidence does not guaranty that the model will produce a faithful answer. Several responses contained errors even when the relevant figure, table, or semantic summary was successfully retrieved: models omitted requested information, misinterpreted relationships in the evidence, reversed explicitly stated associations, or introduced unsupported claims. Table~\ref{tab:evidence-failures} presents representative examples. In all examples, the retrieved evidence contained the information necessary to correctly answer the question, and the point of failure lies within the local verbalization.

\begin{table}[t]
\caption{Examples of incorrect or incomplete answers despite successful retrieval of relevant evidence.}
\label{tab:evidence-failures}
\centering
\small
\begin{tabularx}{\linewidth}{@{}p{0.19\linewidth}p{0.23\linewidth}X@{}}
\toprule
Paper and question & Retrieved evidence & Observed failure \\
\midrule
Multilingual performance package~\cite{kortemeyer2025}: C1, overall image-requirement scores &
Correct Figure~6 summary, ranked first; top 6 &
Apertus returned the text-only and required-image scores (81\% and 49\%) but omitted the unneeded-image score (79\%). The evidence was relevant, but the answer did not cover all requested values. \\

Multilingual performance package~\cite{kortemeyer2025}: C2, overall image-requirement scores &
Correct Figure~6 summary; top 3 &
Qwen correctly returned the values for text-only, required-image and unneeded-image (81\%, 49\% and 79\%), but then incorrectly claimed there are no reported values for required-image per individual subject, only the aggregate. (49\%) it returned. \\

Multilingual performance package~\cite{kortemeyer2025}: C7, AI-versus-Student comparison &
Correct Figure~5 summary, ranked first; top 6 &
LFM correctly stated the general conclusion and then incorrectly claimed that there are no numerical values in the evidence to support the conclusion. \\

Gravitational waves package~\cite{obsgrav2016}: C7, Hanford and Livingston observations &
Figures~1 and~3 retrieved; top 6 &
Apertus reported the correct delay magnitude of approximately 6.9 milliseconds but incorrectly stated that Hanford recorded the signal first. It also omitted the time shift and sign inversion required to explain the alignment. \\

Gravitational waves package~\cite{obsgrav2016}: C8, Hanford and Livingston observations &
Figure~1 retrieved; top 1 &
Qwen correctly returned the delay, time shift, and sign inversion. However the answer reverses the arrival order, contradicting both the paper and its own quoted evidence. \\

Gravitational waves package~\cite{obsgrav2016}: C4, black-hole separation and relative velocity &
Figure~2 retrieved; top 6 &
SmolLM correctly stated that separation decreases and velocity increases, but both axis assignments are reversed. \\
\bottomrule
\end{tabularx}
\end{table}

The examples suggest that correctly retrieving a figure or table establishes that relevant information was available to the model, but does not establish that the model extracted the correct facts, preserved their associations, or addressed every part of the question. The errors observed here range from omissions to incorrect interpretation and unsupported additions. In particular, questions involving multiple values or relationships appear vulnerable to partial answers: a model may reproduce one correct value or trend while failing to provide the remaining information required for a complete response.

The results also suggest that increasing the retrieval depth does not necessarily improve the quality of the answer. In the multilingual performance package, Apertus incorrectly answered the subject-category question C4 (see Table~\ref{tab:conditions-paper1}) with top-six retrieval. For the same question, restricting the evidence to two relevant sentences allowed Apertus to return both the correct categories and values. This contrast is consistent with the possibility that additional, redundant, or distracting evidence can complicate answer generation, although the experiment does not isolate the length of the evidence from all other factors. More retrieved chunks provide a larger evidence set, not a guarantee that the model will select and use the most relevant facts correctly.

Conversely, a small evidence set can be sufficient when it contains the necessary information. In the gravitational waves package, LFM correctly answered the black-hole mass question C3 using the top retrieved chunk, identifying all three central estimates. However, its response to the Hanford--Livingston comparison C8 under a figure-scoped, top-one condition reversed the arrival order despite retrieving Figure~1. Similarly, the minimal-evidence condition for the black-hole separation and velocity question C6 did not consistently produce correct answers across models. Thus, neither increasing nor reducing the number of retrieved chunks yielded uniformly better results across question families and models.

These observations are descriptive rather than causal. The experiment was not designed to isolate retrieval depth, evidence redundancy, or model reasoning as independent variables. Nevertheless, the examples demonstrate an important limitation for the proposed application: retrieval-augmented generation can make relevant document information available to a local model, but additional mechanisms may be needed to ensure that the generated response faithfully reflects that information, see also Section \ref{sec:architecture_outlook}. Evaluating the reader therefore requires assessing not only whether relevant evidence was retrieved, but also whether the answer correctly represents it and covers the question's required elements.

\section{Discussion and Limitations}
\label{sec:discussion}
The proof of concept demonstrates the feasibility of separating document preparation from local, question-driven access to scientific content, but it also exposes limitations in the representation, retrieval, answer generation, and interaction. This section discusses the architectural implications of this separation, the challenges of reliably mediating retrieved evidence through small local models, and the accessibility considerations that remain unaddressed. It concludes by delineating the scope of the current evaluation and the assumptions that require further validation before deployment.

\subsection{The reusable representation is the principal architectural result}

The preparer/reader division allows expensive interpretation to be performed before a reader asks a question. It also creates a persistent object that can be inspected independently of a particular answer model. The worked example in section \ref{sec:worked-example} shows how orienting prose, source-reported values, and visual estimates can coexist in that representation. Future clients could use those resources differently, provided their contracts and provenance remain explicit. Interoperability with such clients is a possibility, not a demonstrated result.

This division does not eliminate model error; it relocates some of it to preparation. A mistaken annotation can be reused consistently across many readers. Institutional preparation is therefore valuable only if it is understood to include responsibility for reviewing and maintaining the representation, rather than merely running a large model once. Structural validation is necessary for reproducible assembly, but cannot substitute for review of object coverage, values, units, approximations, and relationships. The current preparer has no interactive semantic approval gate.

The durable asset is also not yet a standards-conforming enriched EPUB. The directory contract depends on project-specific filenames and validators. A standard EPUB reader receives only the text/card member, and the misleading JSON-LD suffix confers no semantic interoperability. A future format should make the relationship between the standard reading surface and its extensions explicit. The present paper documents this limitation rather than presenting an experimental bundle as a new publication standard.

\subsection{Local language mediation needs a stronger evidence contract}
\label{sec:architecture_outlook}

The diagnostic failures challenge the assumption that selecting a few relevant records leaves only easy wording for a small model. Retrieving the correct evidence is a necessary, but not a sufficient condition, for a correct model response. An adequate answer may require several categories to remain aligned, multiple requested values to be included, and an unsupported causal explanation to be avoided. These are semantic tasks even when no new arithmetic is requested. The current numeric guard checks only whether digit strings occur in the retrieved evidence, and the source links identify objects rather than support for individual claims.

The observations do not justify a general claim that local models are unsuitable, nor do they show that a newer or larger model alone will solve the problem. They show that the evaluated configurations have not demonstrated dependable autonomous Ask-mode operation. Similarly, a perfect score on these nine development conditions would not establish deployment readiness. There is no validated acceptance threshold, and the consequences of errors depend on the task, their detectability, and the interaction through which users encounter them.

One proposed direction is a tiny local model that translates a request into a constrained retrieval query, followed by classical evidence selection and local answer formulation. The latter two stages already exist in elementary form. What is new would be validated query planning and a more deliberate evidence contract: choosing the relevant object or section, identifying required fields and relationships, retaining their qualifications, and selecting a sufficient context within a token budget. Merely lowering \texttt{top\_k} does not implement that contract.

A planner would itself introduce a new failure point, especially when it changes explicit document references. It should therefore be evaluated separately rather than assumed to improve the pipeline. Other proposed options include indexing section evidence, exposing retrieved records directly, deterministic lookup of structured values, and explicit abstention when required evidence is absent. Structured lookup and arithmetic are not features of the current reader. Their discussion here is prospective, not a retrospective reinterpretation of the implementation.

\subsection{Accessible interaction is more than an accurate answer}

The Normal and Structure modes are important because generation should not be the only route to a document. Nevertheless, the presence of these modes does not prove that they provide satisfactory alternatives. Preformatted page extraction can retain reading-order problems; object inspection currently exposes JSON; and no specialized mathematical representation is available. Even an accurate answer may not help a reader efficiently verify a claim or locate its context.

The interface implements keyboard- and screen-reader-oriented mechanisms, but neither conformance nor practical usability has been evaluated. The hard-coded browser language is one concrete defect; the usefulness of focus movement, status announcements, object labels, and the inspection view remains untested. Future work with blind and low-vision readers should examine navigation, evidence checking, error recovery, and the balance between narrative and exact data. Those outcomes cannot be inferred from the model scores, software tests, or our design intentions.

Verification must also be usable nonvisually. Prior work documents the additional work that AI errors can impose on blind users~\cite{misfitting2024}. An Inspect link that leads to a long object record, or access to an original PDF with the same initial barriers, may shift that work rather than resolve it. A future interaction should distinguish source-reported values, visual estimates, model interpretation, and insufficient evidence without requiring the reader to audit the implementation.

\subsection{Limits of this proof of concept}

Our evaluation of local models uses two scientific papers and a small, deliberately diagnostic set of conditions. It does not establish generality across disciplines, visual encodings, layouts, languages, scanned documents, or mathematical content. It does not independently assess whether the preparer found all meaningful objects or correctly reconstructed the plotted data. A strict package can be internally consistent while being incomplete or semantically wrong. Furthermore, the evaluation is for descriptive analysis only, it does not establish causal links.

There is no plain-text question-answering baseline, comparison with conventional reading, ablation, repeated-run estimate, or user study. Evidence-only and lexical modes operate on the same enriched corpus and are not substitutes for such a baseline. We therefore make no causal claim that the semantic layer improves accuracy or reduces effort. A compact follow-up can first address evidence fidelity and retain request-level traces before a broader comparison is warranted; the present proof of concept need not be mistaken for that later study.

Finally, the intended resource and privacy advantages remain conditional. CPU-only operation is implemented, but device-specific feasibility is not characterized. Reading-time locality does not remove preparation-time data disclosure, software supply-chain considerations, or the need to obtain permission to distribute source PDFs and derived packages. These are deployment requirements, not benefits established by the current experiment.

\section{Conclusion}

The proposed architecture separates institution-side interpretation of a scientific PDF from local exploration of a persistent prepared representation. It combines an EPUB text/card surface with project-specific semantic and retrieval resources and a browser reader that offers ordinary reading, object inspection, and optional local question answering. The worked example in section \ref{sec:worked-example} demonstrates preservation of descriptions and numerical annotations within a strict package contract.

The preliminary proof-of-concept within the reader also exposes a clear boundary: making relevant evidence available does not make small-model answers dependable. Recorded failures include omissions, incorrect relationships, missing section coverage, context overflow, and timeouts. The resulting 4--13 point totals out of 18 describe this diagnostic exercise, not the general quality of the model or the accessibility measured. The next architectural question is how to deliver sufficient, coherent, and verifiable evidence to a narrowly constrained local language interface while preserving non-generative access. Conventional accessible publishing and evaluation with intended users remain essential; neither is replaced by the preparer, the package validator, or the language model.

\section*{Artifact and Data Availability}

The prepared packages for both articles~\cite{kortemeyer2025, obsgrav2016} used within this work, the raw evaluation data of our proof-of-concept on which section \ref{sec:results} is based, and code for both the preparer and the reader are open source and available on GitLab~\cite{pr_repo}. Contact the authors in case of questions.

\begin{acks}
This work has been funded in part by swissuniversities PgB Chancengleichheit. 
\end{acks}
\bibliographystyle{ACM-Reference-Format}
\bibliography{references}
\appendix
\section{Diagnostic Questions and Reference Answers}
\label{app:questions}

This section provides the reference answers and accepted deviations used to evaluate the model responses. The evidence scopes and retrieval settings for each condition are specified in Tables~\ref{tab:conditions-paper1} and~\ref{tab:conditions-paper2}, respectively. The reference answers are not intended as exact-match targets; the acceptance criteria specify which information is required for full or partial credit and which additional information is permitted.

\subsection{Multilingual Performance of a Multimodal Artificial Intelligence System on Multisubject Physics Concept Inventories}

The questions for the multilingual performance package~\cite{kortemeyer2025} are grouped into four question families. Conditions within the same family share the same question language, but may differ in their scope of evidence or retrieval settings.
See Table \ref{tab:answers-paper1} for reference answers and accepted deviations.

\begin{table}[p]
\caption{Questions, reference answers, and acceptance criteria for the multilingual performance package.}
\label{tab:answers-paper1}
\centering
\small
\begin{tabularx}{\linewidth}{@{}p{0.10\linewidth}p{0.32\linewidth}X@{}}
\toprule
Condition & Reference answer & Acceptance criteria \\
\midrule
C1--C3 &
``The overall text-only, unneeded-image, and required-image scores are 81\%, 79\%, and 49\%, respectively.'' &
All three values must be present for full credit. Scores may be named explicitly, and additional information is permitted provided it is correct and does not cause confusion. \\
\addlinespace
C4--C6 &
``Thermodynamics: 85.2\%; Laboratory Skills: 35.0\%.'' &
Both categories and their corresponding values are required for full credit. Additional information is permitted if correct and unambiguous. \\
\addlinespace
C7--C8 &
``AI outperformed students in 68.9\% of cases. This is based on comparisons of inventories within evaluated languages with published, average student scores after tutorings.'' &
The numerical value may be omitted provided the central conclusion that AI outperformed students is correct. \\
\addlinespace
C9 &
``The study has methodological limitations, including translation quality, image interpretation, and prompting strategies, as well as the lack of evaluation of physics reasoning. These limitations warrant a cautious interpretation of its findings.'' &
Answers may be more extensive and provide further detail, provided the information is correct and does not cause confusion. \\
\bottomrule
\end{tabularx}
\end{table}

\subsection{Observation of Gravitational Waves from a Binary Black Hole Merger}

The questions for the gravitational waves package~\cite{obsgrav2016} concern the reported observations and analysis of the gravitational-wave event. As in the multilingual performance package, conditions sharing a question family use identical question wording, but differ in their evidence-selection settings.
See Table \ref{tab:answers-paper2} for reference answers and accepted deviations.

\begin{table}[p]
\caption{Questions, reference answers, and acceptance criteria for the gravitational waves package.}
\label{tab:answers-paper2}
\centering
\small
\begin{tabularx}{\linewidth}{@{}p{0.10\linewidth}p{0.35\linewidth}X@{}}
\toprule
Condition & Reference answer & Acceptance criteria \\
\midrule
C1--C3 &
``The primary, secondary, and final black holes had source-frame masses of 36, 29, and 62 solar masses, respectively.'' &
All three values must be correctly associated with their respective objects for full credit. Listing the values in the correct order is sufficient; uncertainties are optional. Source-frame and detector-frame masses must not be confused. Correctly identifying only one or two masses earns partial credit. Additional information is permitted if correct, relevant, and unambiguous. \\
\addlinespace
C4--C6 &
``The left-hand axis shows effective relative velocity, which increases towards the merger. The right-hand axis shows effective black-hole separation, which decreases towards the merger.'' &
Full credit requires both correct axis assignments and both trends. Units, curve colours, and approximate numerical values are optional. Correct trends without axis assignments, or correct assignments without trends, earn partial credit. Additional information is permitted if correct, relevant, and unambiguous. \\
\addlinespace
C7--C8 &
``The signal arrived first at Livingston, followed approximately 6.9 milliseconds later by Hanford. The Hanford trace was shifted in time to align the arrivals and inverted in sign to account for the detectors' relative orientations.'' &
Full credit requires the correct arrival order, approximate delay, time shift, and sign inversion. L1 and H1 are acceptable alternatives for Livingston and Hanford, respectively; approximately 7 milliseconds is an acceptable delay. The delay uncertainty and explanation of the sign inversion are optional. Correct arrival order and delay without the inversion earn partial credit. Additional information is permitted if correct, relevant, and unambiguous. \\
\addlinespace
C9 &
``The authors compared the final black hole's mass and spin inferred independently from the early and late stages of the coalescence and found them consistent. They tested for deviations from general relativity in the coefficients describing the inspiral phase of the gravitational waveform and found no evidence of deviations. Finally, assuming a modified gravitational-wave dispersion relation, they constrained the graviton's Compton wavelength, thereby obtaining an upper bound on its mass. All three tests were consistent with general relativity.'' &
Full credit requires all three tests and their overall conclusion, regardless of presentation or order. Equations and numerical bounds are optional. The third test may be described in terms of gravitational-wave dispersion or a constraint on the graviton mass. If numerical bounds are given, they must be consistent with a Compton wavelength greater than $10^{13}$ km and a graviton mass below $1.2 \times 10^{-22}$ eV/$c^2$, under the stated assumption. These are bounds, not measurements of a nonzero graviton mass. Covering only one or two tests, or stating only the overall conclusion, earns partial credit. Claims of a detected violation of general relativity or proof of its universal correctness are unacceptable. \\
\bottomrule
\end{tabularx}
\end{table}

\section{Condition-Level Scores}
\label{app:scores}
\begin{table}[t]
\caption{Original condition-level scores for the multilingual performance package.}
\label{tab:scores_p1}
\centering
\begin{tabular}{@{}lrrrrrrrrrr@{}}
\toprule
Model & C1 & C2 & C3 & C4 & C5 & C6 & C7 & C8 & C9 & Total \\
\midrule
Apertus 1.5B & 1 & 0 & 0 & 0 & 0 & 2 & 0 & 1 & 0 & 4 \\
Qwen3.5 0.8B & 2 & 1 & 0 & 2 & 2 & 1 & 0 & 0 & 0 & 8 \\
SmolLM2-1.7B-Instruct & 2 & 2 & 2 & 2 & 2 & 2 & 0 & 1 & 0 & 13 \\
LFM2.5-1.2B-Instruct & 2 & 2 & 0 & 0 & 0 & 0 & 0 & 1 & 0 & 5 \\
\bottomrule
\end{tabular}
\end{table}

\begin{table}[t]
\caption{Original condition-level scores for the gravitational waves package.}
\label{tab:scores_p2}
\centering
\begin{tabular}{@{}lrrrrrrrrrr@{}}
\toprule
Model & C1 & C2 & C3 & C4 & C5 & C6 & C7 & C8 & C9 & Total \\
\midrule
Apertus 1.5B & 1 & 1 & 1 & 0 & 0 & 1 & 1 & 0 & 0 & 5 \\
Qwen3.5 0.8B & 0 & 0 & 0 & 2 & 2 & 1 & 2 & 1 & 0 & 8 \\
SmolLM2-1.7B-Instruct & 0 & 2 & 0 & 1 & 2 & 1 & 0 & 1 & 0 & 7 \\
LFM2.5-1.2B-Instruct & 2 & 1 & 2 & 2 & 1 & 1 & 2 & 1 & 0 & 12 \\
\bottomrule
\end{tabular}
\end{table}

Tables \ref{tab:scores_p1} and \ref{tab:scores_p2} retain all original scores per package and per question. 

\end{document}